\pdfoutput=1
\documentclass[11pt]{article}
\usepackage[preprint]{acl}
\usepackage{amsmath}
\usepackage{bm}
\usepackage{booktabs}
\usepackage[T1]{fontenc}
\usepackage{graphicx}
\usepackage{inconsolata}
\usepackage[utf8]{inputenc}
\usepackage{latexsym}
\usepackage{microtype}
\usepackage{subcaption}
\usepackage{tcolorbox}
\usepackage{times}
\title{Retrieval of Temporal Event Sequences from Textual Descriptions}

\author{%
    Zefang Liu\thanks{These authors contributed equally to this work.} \\
    Georgia Institute of Technology \\
    Atlanta, GA 30332, USA \\
    \texttt{liuzefang@gatech.edu} \\\And
    Yinzhu Quan$^*$ \\
    Georgia Institute of Technology \\
    Atlanta, GA 30332, USA \\
    \texttt{yquan9@gatech.edu} \\
}
\begin{document}
\maketitle
\begin{abstract}
Retrieving temporal event sequences from textual descriptions is crucial for applications such as analyzing e-commerce behavior, monitoring social media activities, and tracking criminal incidents. To advance this task, we introduce TESRBench, a comprehensive benchmark for temporal event sequence retrieval (TESR) from textual descriptions. TESRBench includes diverse real-world datasets with synthesized and reviewed textual descriptions, providing a strong foundation for evaluating retrieval performance and addressing challenges in this domain. Building on this benchmark, we propose TPP-Embedding, a novel model for embedding and retrieving event sequences. The model leverages the TPP-LLM framework, integrating large language models (LLMs) with temporal point processes (TPPs) to encode both event texts and times. By pooling representations and applying a contrastive loss, it unifies temporal dynamics and event semantics in a shared embedding space, aligning sequence-level embeddings of event sequences and their descriptions. TPP-Embedding demonstrates superior performance over baseline models across TESRBench datasets, establishing it as a powerful solution for the temporal event sequence retrieval task.
\end{abstract}
\section{Introduction}

Temporal event sequence retrieval \citep{gupta2022learning} plays a crucial role in various applications, such as e-commerce user activity analysis, social media monitoring, and crime tracking. These sequences combine temporal information with event types, making them more complex than traditional text data. Effective retrieval requires models capable of capturing both time-sensitive dynamics and structured relationships within the sequences. While traditional language models perform well for general text retrieval \citep{kashyap2024comprehensive}, they often struggle to handle the unique temporal and structural complexities of event sequences.

\begin{figure}[t!]
    \centering
    \includegraphics[width=\linewidth]{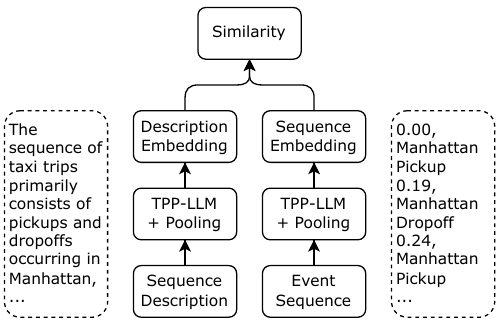}
    \caption{TPP-Embedding framework with one TESRBench example, where the model embeds both textual descriptions and temporal event sequences using a shared TPP-LLM framework, applies pooling to generate fixed-length representations, and uses contrastive learning with similarity scores to align matching pairs for effective event sequence retrieval.}
    \label{fig:tpp-embedding}
\end{figure}

To address these challenges, we introduce TESRBench\footnote{Datasets: \url{https://huggingface.co/tppllm}}, a comprehensive benchmark for evaluating temporal event sequence retrieval (TESR) from textual descriptions. TESRBench comprises diverse real-world event sequence datasets with synthesized and reviewed textual descriptions, offering a strong foundation for benchmarking retrieval models. It highlights the complexities of aligning event sequences with textual descriptions and provides a standardized platform for evaluating model performance, uncovering key challenges, and identifying opportunities for improvement in temporal and contextual modeling.

Building on this benchmark, we propose TPP-Embedding\footnote{GitHub repository: \url{https://github.com/zefang-liu/TPP-Embedding}}, a novel framework for temporal event sequence retrieval that extends the TPP-LLM model \citep{liu2024tppllmm}. TPP-LLM integrates temporal encoding for event times and textual embeddings for event types within a large language model (LLM) backbone to model temporal point processes (TPPs). Extending this framework, TPP-Embedding aligns sequence-level representations of event sequences and their textual descriptions in a shared embedding space. By modeling the interdependencies between events and their temporal context, TPP-Embedding generates richer, contextually informed embeddings optimized for retrieval tasks. Evaluated across TESRBench datasets, TPP-Embedding demonstrates superior performance over text-based baselines and generalizes effectively across different event domains.

In this paper, our key contributions are: (1) Introducing TESRBench, a benchmark for evaluating TESR models with diverse datasets; (2) Proposing TPP-Embedding, which integrates temporal and event-type information for accurate event sequence retrieval from descriptions; and (3) Showcasing the scalability and flexibility of our approach through multi-domain experiments.

\section{Related Work}

Recent developments in sentence representation models, such as Sentence-BERT \citep{reimers2019sentence}, have significantly improved retrieval tasks by enabling efficient semantic similarity searches using transformer-based embeddings \citep{vaswani2017attention}. While these models perform well in standard text retrieval tasks \citep{lin2022pretrained}, they struggle with the temporal and event-specific complexities of event sequence data. To address these challenges, temporal point process (TPP) models \citep{mei2017neural,shchur2021neural,xue2023easytpp} have been adapted for retrieval tasks. NeuroSeqRet \citep{gupta2022learning,gupta2023retrieving} introduces a neural framework for continuous-time event sequence retrieval by leveraging marked TPPs to model temporal dynamics and using a trainable unwarping function, neural relevance models, and hashing techniques to optimize retrieval efficiency. However, despite these advancements, existing models either treat event types as categorical inputs, limiting their ability to capture rich event semantics, or treat entire sequences as text, ignoring their temporal dependencies. 

Recently, \citet{liu2024tppllmm} proposed TPP-LLM, a framework that integrates large language models (LLMs) with TPPs to capture event semantics and temporal dynamics for event sequence modeling and prediction. While TPP-LLM focuses on predicting future event types and times using both textual and temporal information, our proposed TPP-Embedding extends this framework to the task of retrieving temporal event sequences from textual descriptions. By introducing a shared embedding space for sequences and descriptions and employing contrastive learning, our model effectively aligns sequence-level representations with natural language descriptions, enabling retrieval while maintaining temporal and semantic dependencies.

\section{Benchmark}
\label{sec:datasets}

In this section, we present TESRBench, a comprehensive benchmark designed to evaluate temporal event sequence retrieval (TESR) from textual descriptions. We provide an overview of its key components, including detailed dataset summaries, the methodology for generating event sequence descriptions, and the evaluation process used to assess the quality of these descriptions.

\subsection{Dataset Summaries}
\label{sec:dataset-summaries}

\begin{table*}[!h]
\centering
\small
\begin{tabular}{lrrrrrr}
\toprule
\textbf{Dataset} & \textbf{Domain} & \textbf{\# of Types} & \textbf{\# of Events} & \textbf{\# of Seq.} & \textbf{Avg. Seq. Length} & \textbf{Time Unit} \\
\midrule
Stack Overflow & Social Networks & 25 & 187,836 & 3,336 & 56.31 & Month \\
Chicago Crime & Urban Dynamics & 20 & 202,333 & 4,033 & 50.17 & Month \\
NYC Taxi Trip & Transportation & 8 & 362,374 & 2,957 & 122.55 & Hour \\
U.S. Earthquake & Natural Disasters & 3 & 29,521 & 3,009 & 9.81 & Day  \\
Amazon Review & E-Commerce & 18 & 127,054 & 2,245 & 56.59 & Week \\
\bottomrule 
\end{tabular}
\caption{Dataset statistics overview of event sequences in TESRBench. (\# = Number.)}
\label{tab:datasets}
\end{table*}

TESRBench is built on five real-world datasets from diverse domains: Stack Overflow, Chicago Crime, NYC Taxi Trip, U.S. Earthquake, and Amazon Review. Each dataset captures sequences of event-based information within specific time periods but lacks accompanying textual sequence descriptions. To address this, we generate textual descriptions for these event sequences using GPT-4o-mini \citep{achiam2023gpt}, creating objective summaries that emphasize the order and timing of events while preserving their essential structure. Details of the description generation and evaluation processes are provided in subsequent subsections. Examples of the data from TESRBench are included in Appendix \ref{sec:data-examples} for further reference.

The datasets in TESRBench span various domains and offer rich opportunities for analysis. Table \ref{tab:datasets} presents an overview of their key statistics, using the same train/validation/test splits as \citet{liu2024tppllmm}. The \textbf{Stack Overflow} \citep{stackexchange2024dump} dataset tracks non-tag-related badges earned between January 2022 and December 2023, comprising 3,336 sequences across 25 event types. The \textbf{Chicago Crime} \citep{chicago2024crimes} dataset focuses on the top 20 crime types and blocks with 30-120 incidents during the same time period, yielding 4,033 sequences across 20 crime categories. The \textbf{NYC Taxi Trip} \citep{nyctaxi2024trips} dataset captures trips from May 1-7, 2013, excluding Staten Island, with 2,957 sequences across 8 location categories. The \textbf{U.S. Earthquake} \citep{usgs2024earthquakes} dataset records 3,009 sequences of earthquake events from January 2020 to December 2023, categorized into 3 magnitude levels. Finally, the \textbf{Amazon Review} \citep{ni2019justifying} dataset comprises 2,245 sequences of 40-200 reviews per user between January and June 2018, spanning 18 categories. Collectively, these datasets establish a robust foundation for evaluating models on diverse temporal event sequence retrieval tasks.

\subsection{Description Generation}
\label{sec:description-generation}

To create textual descriptions for the event sequences in TESRBench, we employ a structured process using GPT-4o-mini \citep{achiam2023gpt}. The process begins with crafting a system message, as illustrated in Figure \ref{fig:system-message}, which guides GPT-4o-mini to produce objective summaries that focus on the order and timing of events. The instructions explicitly avoid interpreting behaviors or including specific numbers or timestamps, ensuring consistency and objectivity in the generated summaries. For each dataset, specific prompts are designed to reflect the context of the event sequences, as detailed in Table \ref{tab:user-messages}. These prompts present sequences of events with timestamps and event types, formatted to highlight the unique characteristics of each dataset. GPT-4o-mini processes these prompts and generates concise textual descriptions that capture key patterns and trends, providing an accurate summary of how events unfold over time. This approach ensures that the generated descriptions are well-aligned with the underlying temporal and contextual dynamics of the event sequences.

\begin{figure}[!h]
  \centering
  \begin{tcolorbox}
    \textbf{System Message:}

    You are an expert in summarizing event sequences. Your task is to provide a 2-5 sentence objective summary of the sequence's key patterns and trends without interpreting any behaviors or motivations. Focus on the sequence's order and timing, emphasizing how the events unfold over time. Describe general trends such as whether certain event types occur earlier or later, or if events cluster in certain periods. Avoid including exact numbers or timestamps.
  \end{tcolorbox}
  \caption{Instructions for generating objective summaries of event sequences, focusing on the order, timing, and general trends without including specific numbers or timestamps.}
  \label{fig:system-message}
\end{figure}

\begin{table*}[!h]
\centering
\small
\begin{tabular}{p{.15\linewidth}p{.75\linewidth}}
\toprule
\textbf{Dataset} & \textbf{Description} \\
\midrule
Stack Overflow & 
Here is a sequence of badges earned by a user on Stack Overflow, with relative timestamps (in months) and badge names. Please provide a summary that describes the timing and order of events:
\newline
\newline
\{event\_sequence\} \\
\midrule
Chicago Crime & 

Here is a sequence of crime incidents reported at a block in Chicago, with relative timestamps (in months) and crime types. Please provide a summary that describes the timing and order of events:
\newline
\newline
\{event\_sequence\}\\
\midrule
NYC Taxi Trip &

Here is a sequence of taxi trips taken by a driver in New York City, with relative timestamps (in hours) and trip locations. Please provide a summary that describes the timing and order of events:
\newline
\newline
\{event\_sequence\}\\
\midrule
U.S. Earthquake & 

Here is a sequence of earthquake events in the U.S., with relative timestamps (in days) and magnitude categories. Please provide a summary that describes the timing and order of events:
\newline
\newline
\{event\_sequence\}\\
\midrule
Amazon Review & 

Here is a sequence of product reviews submitted by a user on Amazon, with relative timestamps (in weeks) and review categories. Please provide a summary that describes the timing and order of events:
\newline
\newline
\{event\_sequence\}\\
\bottomrule
\end{tabular}
\caption{Overview of dataset-specific prompts, describing event sequences from various domains.}
\label{tab:user-messages}
\end{table*}

\subsection{Description Evaluation}
\label{sec:description-evaluation}

To assess the quality of the generated descriptions for temporal event sequences, we define a set of evaluation criteria and scoring scales. Using GPT-4o and GPT-4o-mini, these criteria measure the descriptions across five dimensions: accuracy, coverage, fidelity, clarity, and conciseness. The definitions of these criteria, along with their corresponding scoring scales, are as follows:
\textbf{Accuracy}: Does the description correctly represent the sequence of events, focusing on the event types, their order, and timing? (1 = Completely inaccurate, 5 = Completely accurate)
\textbf{Coverage}: Does the description include all significant events and key details of the sequence, without omitting critical information? (1 = Very incomplete, 5 = Fully comprehensive)
\textbf{Fidelity}: To what extent does the description capture and reflect the temporal relationships and patterns (e.g., clustering, trends, or intervals) in the event sequence? (1 = No temporal fidelity, 5 = High temporal fidelity)
\textbf{Clarity}: Is the description easy to understand, with clear language and a logical structure that aids comprehension? (1 = Very unclear, 5 = Very clear)
\textbf{Conciseness}: Does the description provide the necessary information in a succinct manner, avoiding unnecessary verbosity or redundancy? (1 = Overly verbose or incomplete, 5 = Very concise and complete)

\begin{table}[h]
    \centering
    \small
    \begin{tabular}{lccccc}
        \toprule
        \textbf{Dataset} & \textbf{Acc.} & \textbf{Cov.} & \textbf{Fid.} & \textbf{Cla.} & \textbf{Con.} \\
        \midrule
        StackOverflow & 4.10 & 4.05 & 4.25 & 4.94 & 4.56 \\
        Crime & 4.01 & 4.00 & 4.18 & 4.98 & 4.67 \\
        Taxi & 4.44 & 4.03 & 4.46 & 4.89 & 4.36 \\
        Earthquake & 4.36 & 4.31 & 4.42 & 4.96 & 4.95 \\
        Amazon & 4.66 & 4.33 & 4.74 & 4.99 & 4.82 \\
        \bottomrule
    \end{tabular}
    \caption{Evaluation scores of descriptions generated for event sequences across different datasets in TESRBench from GPT-4o. 
    (Acc. = Accuracy, Cov. = Coverage, Fid. = Fidelity, Cla. = Clarity, Con. = Conciseness.)}
    \label{tab:evaluation-scores-gpt-4o}
\end{table}

\begin{table}[h]
    \centering
    \small
    \begin{tabular}{lccccc}
        \toprule
        \textbf{Dataset} & \textbf{Acc.} & \textbf{Cov.} & \textbf{Fid.} & \textbf{Cla.} & \textbf{Con.} \\
        \midrule
        StackOverflow & 5.00 & 5.00 & 5.00 & 5.00 & 5.00 \\
        Crime  & 5.00 & 4.87 & 4.96 & 5.00 & 4.86 \\
        Taxi & 5.00 & 4.99 & 4.99 & 5.00 & 4.99 \\
        Earthquake & 4.99 & 4.99 & 4.85 & 5.00 & 4.93 \\
        Amazon & 5.00 & 5.00 & 4.98 & 5.00 & 5.00 \\
        \bottomrule
    \end{tabular}
    \caption{Evaluation scores of descriptions generated for event sequences across different datasets in TESRBench from GPT-4o-mini. 
    (Acc. = Accuracy, Cov. = Coverage, Fid. = Fidelity, Cla. = Clarity, Con. = Conciseness.)}
    \label{tab:evaluation-scores-gpt-4o-mini}
\end{table}

The averaged evaluation scores across datasets are presented in Tables \ref{tab:evaluation-scores-gpt-4o} and \ref{tab:evaluation-scores-gpt-4o-mini}, which show the evaluation of descriptions generated by GPT-4o-mini using two evaluators: GPT-4o and GPT-4o-mini. GPT-4o’s evaluation scores indicate strong performance, particularly in clarity and conciseness, though with slightly lower scores in accuracy, coverage, and fidelity compared to GPT-4o-mini’s evaluation. On the other hand, GPT-4o-mini’s evaluation highlights consistently high scores across all dimensions, reflecting its alignment with the generated descriptions. The complementary insights provided by both evaluations emphasize the overall effectiveness of the descriptions in summarizing event sequences, while also preserving the temporal and contextual relationships critical to the datasets. These results underline the robustness of the generated descriptions when assessed using different evaluators.

\section{Methodology}

In this section, we introduce TPP-Embedding, an extension of TPP-LLM \citep{liu2024tppllmm}, designed to embed both event sequences and textual descriptions into a shared embedding space, enabling effective retrieval based on similarity.

\subsection{Model Architecture}

Given a set of textual descriptions $\mathcal{D} = \{d_1, d_2, \dots, d_m\}$ and a set of temporal event sequences $\mathcal{S} = \{s_1, s_2, \dots, s_n\}$, the task is to retrieve the most relevant sequence $s^* \in \mathcal{S}$ for a given description $d_j$. Each event sequence $s_i$ consists of a series of events $\{e_{i,1}, e_{i,2}, \dots, e_{i,n_i}\}$, where each event $e_{i,j}$ is represented by an event time $t_{i,j}$ and an event type $k_{i,j}$. Thus, the sequence can be written as $s_i = \{(t_{i,1}, k_{i,1}), (t_{i,2}, k_{i,2}), \dots, (t_{i,n_i}, k_{i,n_i})\}$. The goal is to embed both descriptions $d_j$ and event sequences $s_i$ into a shared embedding space for effective retrieval.

\begin{figure}[!h]
    \centering
    \includegraphics[width=\linewidth]{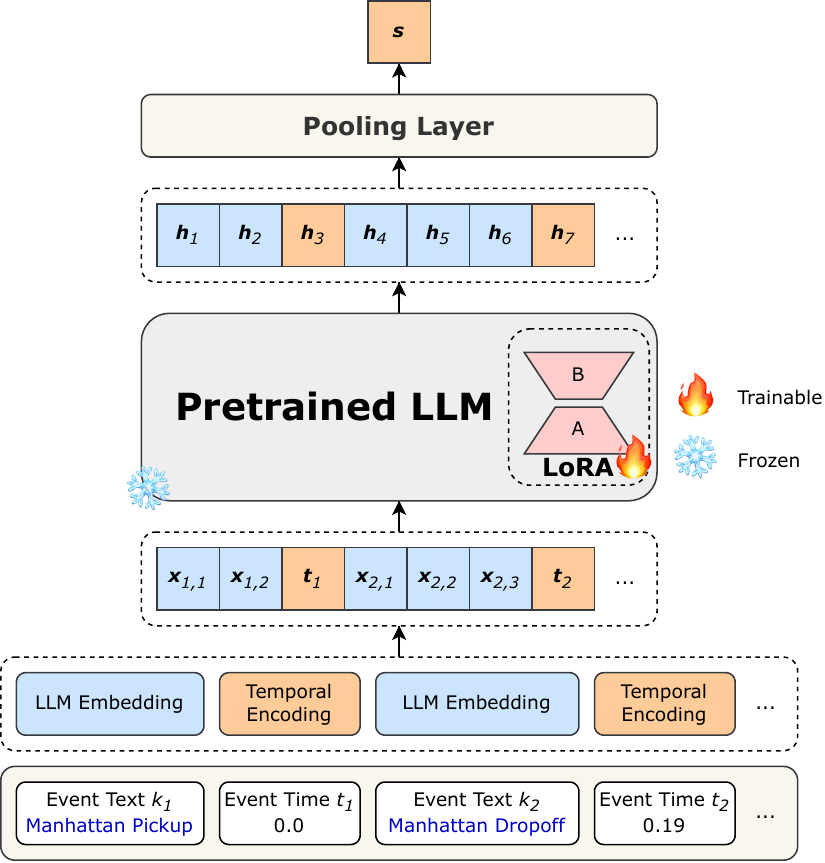}
    \caption{TPP-Embedding architecture, illustrating the embedding process for a event sequence through the integration of temporal and text representations, followed by processing with a large language model and a pooling layer to generate a fixed-length sequence representation.}
    \label{fig:tpp-embedding-model}
\end{figure}

\textbf{Embedding Event Sequences.} As illustrated by Figure \ref{fig:tpp-embedding-model}, TPP-Embedding builds upon TPP-LLM \citep{liu2024tppllmm} by embedding event sequences through the integration of temporal and event-type representations. For each event $e_{i,j}$, the temporal embedding is computed as $\bm{t}_{i,j} = f_t(t_{i,j})$, where $f_t$ is a temporal encoding function \citep{zhang2020self,zuo2020transformer}. Each event type text $k_{i,j}$ is tokenized by the large language model (LLM) tokenizer and embedded using its embedding layer, resulting in $\bm{X}_{i,j} = [\bm{x}_{i,j,1}, \bm{x}_{i,j,2}, \dots, \bm{x}_{i,j,n_j}]$. The temporal and type embeddings are concatenated to form the final event representation $\bm{E}_{i,j}$. These event embeddings are then passed through the LLM to obtain hidden states $\bm{H}_i = [\bm{h}_{i,1}, \bm{h}_{i,2}, \dots, \bm{h}_{i,l_i}] = \text{LLM}([\bm{E}_{i,1}, \bm{E}_{i,2}, \dots, \bm{E}_{i,n_i}])$. Finally, a pooling operation \citep{reimers2019sentence} is applied to produce a fixed-length representation of the sequence: $\bm{s}_i = \text{Pool}(\bm{H}_i)$.

\textbf{Embedding Descriptions.} Textual descriptions $d_j$ are embedded using the same LLM and tokenizer as the event sequences. The description is tokenized and passed through the LLM, resulting in hidden states. A pooling operation is then applied to obtain the final description embedding: $\bm{d}_j = \text{Pool}(\text{LLM}(d_j))$. By embedding descriptions and sequences in the same space, TPP-Embedding enables retrieval based on their similarity.

\subsection{Training Objective}

To align the embeddings of descriptions and their corresponding event sequences, we employ a contrastive learning framework. Positive pairs $(d_i, s_i)$ consist of a description and its matching event sequence, while other sequences in the batch serve as negatives. The cosine similarity between description and sequence embeddings is computed as $\text{sim}(d_i, s_j) = \frac{\bm{d}_i \cdot \bm{s}_j}{\|\bm{d}_i\| \|\bm{s}_j\|}$. The training objective uses a multiple negatives ranking loss \citep{henderson2017efficient} to maximize similarity for positive pairs and minimize it for negative pairs. The loss function is given by:
\begin{equation}
    \mathcal{L} = - \log \frac{\exp(\text{sim}(d_i, s_i))}{\sum_{j} \exp(\text{sim}(d_i, s_j))} .
\end{equation}
This encourages the model to rank the correct event sequence higher than incorrect ones for each description. To improve efficiency, we apply 4-bit precision quantization \citep{dettmers2024qlora} to reduce memory usage and use low-rank adaptation (LoRA) \citep{hu2021lora} to fine-tune a small subset of parameters while keeping the rest frozen. These enhancements allow for efficient fine-tuning and deployment without compromising retrieval performance.

\section{Experiments}

In this section, we present a detailed overview of the baseline models used for comparison, the evaluation metrics employed, the experimental setup, the results obtained, and the ablation studies conducted.

\subsection{Baselines}
To enable evaluation with common embedding models, we transform temporal event sequences into a textual format by concatenating events within a sequence. Each event is represented by its relative timestamp followed by the corresponding event type text, separated by a comma. These events are concatenated with line breaks, resulting in a single textual representation for each event sequence. This approach ensures that the temporal and semantic information is preserved for text-based embeddings.

We compare TPP-Embedding against several widely used embedding models: All-MiniLM-L12-v2 \citep{wang2020minilm}, All-MPNet-Base-v2  \citep{song2020mpnet}, BGE-Large-En-v1.5 \citep{xiao2023cpack}, MxbAI-Embed-Large-v1 \citep{li2023angle,emb2024mxbai}, and Multilingual-E5-Large-Instruct \citep{wang2024multilingual}. These models are designed for generating sentence embeddings and are adapted here for retrieving the most relevant event sequences based on descriptions.

To ensure a fair comparison, all baseline models are fine-tuned using a contrastive learning framework. Specifically, we employ the multiple negatives ranking loss \cite{henderson2017efficient}, which treats a description and its corresponding event sequence as a positive pair, while all other mismatched pairs within the batch are considered negatives. This fine-tuning process aligns the embeddings of matching descriptions and sequences while separating non-matching ones. In addition, Table \ref{tab:model-sizes} provides an overview of the total parameters and trainable parameters for each baseline model. While the baseline models require fine-tuning all parameters, TPP-Embedding leverages LoRA for efficient fine-tuning.

\begin{table}[!h]
\centering
\small
\begin{tabular}{lcc}
\toprule
\textbf{Model} & \textbf{Parameters} & \textbf{Trainable} \\
\midrule
MiniLM-L12 & 33.4M & 33.4M \\
MPNet-Base & 109M & 109M \\
BGE-Large & 335M & 335M \\
MxbAI-Large & 335M & 335M \\
mE5-Large & 560M & 560M \\
\textbf{TPP-Llama} & 1.1B & 4.6M \\
\textbf{TPP-Llama-Chat} & 1.1B & 4.6M \\
\bottomrule
\end{tabular}
\caption{Numbers of total and trainable model parameters. (M = Million, B = Billion.)}
\label{tab:model-sizes}
\end{table}

\subsection{Evaluation Metrics}

The temporal event sequence and description matching task is framed as a retrieval problem, where the model retrieves the correct event sequence for each description by ranking all event sequences based on their similarity to the description embeddings. We evaluate retrieval quality using two metrics: Mean Reciprocal Rank (MRR) and Recall@K. MRR measures the ranking position of the correct sequence, providing an average of reciprocal ranks across all queries, while Recall@K calculates the proportion of cases where the correct sequence is included in the top K results.

\subsection{Experimental Setups}

For the baseline models, we use the AdamW optimizer \citep{loshchilov2017decoupled}, training for 15 epochs with a learning rate of 2e-5, a cosine scheduler, a warmup ratio of 0.1, and a batch size of 8. TPP-Embedding integrates temporal positional encoding for event times \citep{zuo2020transformer}, with event type embeddings placed before the temporal embedding \citep{liu2024tppllmm}. Two foundation models are employed: TinyLlama-1.1B-Intermediate-Step-1431k-3T (TPP-Llama) and TinyLlama-1.1B-Chat-v1.0 \citep{zhang2024tinyllama} (TPP-Llama-Chat). We utilize all hidden states with mean pooling \citep{reimers2019sentence} and apply 4-bit quantization \citep{dettmers2024qlora}. LoRA \citep{hu2021lora} is used with a rank of 16 and dropout of 0.05, targeting the attention projection matrices. The model is trained for 25 epochs with a learning rate of 4e-4, a cosine scheduler, a warmup ratio of 0.02, and a batch size of 8. All experiments are conducted five times, with average results reported. The experiments were run on NVIDIA A100 and H100 GPUs. Additional experimental setups are provided in Appendix \ref{sec:more-setup}.

\subsection{Experimental Results}

\begin{table*}[!h]
\centering
\small
\begin{tabular}{cccccc}
\toprule
\textbf{Model} & \textbf{StackOverflow} & \textbf{Crime} & \textbf{Taxi} & \textbf{Earthquake} & \textbf{Amazon} \\
\midrule
MiniLM-L12 & 0.501 / 0.695 & 0.808 / 0.931 & 0.159 / 0.239 & 0.676 / 0.895 & 0.459 / 0.573 \\
MPNet-Base & 0.620 / 0.775 & 0.924 / 0.980 & 0.246 / 0.364 & 0.733 / 0.923 & \textbf{0.665} / 0.756 \\
BGE-Large & 0.632 / 0.786 & 0.922 / 0.985 & 0.286 / 0.415 & 0.736 / 0.928 & \underline{0.656} / 0.746 \\
MxbAI-Large & 0.627 / 0.782 & 0.924 / 0.982 & 0.271 / 0.426 & 0.717 / 0.914 & 0.650 / 0.747 \\
mE5-Large & 0.658 / 0.804 & 0.941 / 0.987 & 0.261 / 0.389 & 0.748 / 0.921 & 0.617 / 0.716 \\
\textbf{TPP-Llama} & \textbf{0.741} / \textbf{0.880} & \underline{0.958} / \underline{0.992} & \underline{0.468} / \underline{0.680} & \textbf{0.760} / \underline{0.946} & 0.641 / \underline{0.763} \\
\textbf{TPP-Llama-Chat} & \underline{0.729} / \underline{0.865} & \textbf{0.961} / \textbf{0.994} & \textbf{0.475} / \textbf{0.691} & \underline{0.759} / \textbf{0.953} & 0.646 / \textbf{0.767} \\
\bottomrule
\end{tabular}
\caption{Comparison of average MRR and Recall@5 across TESRBench datasets in event sequence retrieval.}
\label{tab:results}
\end{table*}

\begin{figure*}[!h]
\centering
\begin{subfigure}[b]{0.195\textwidth}
    \centering
    \includegraphics[width=\textwidth]{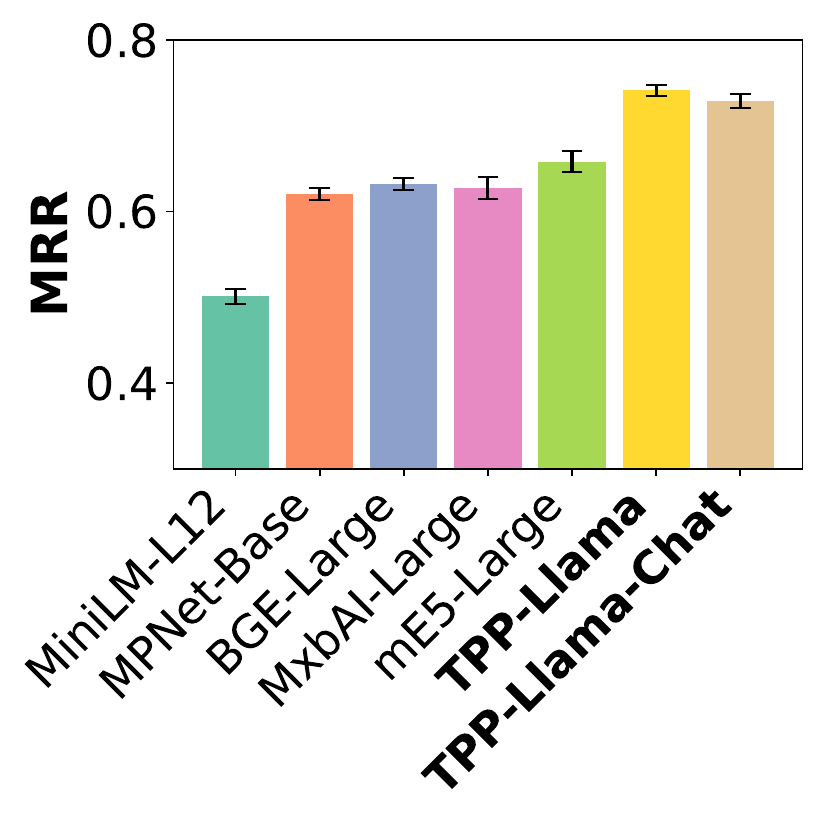}
    \caption{Stack Overflow}
\end{subfigure}
\hfill
\begin{subfigure}[b]{0.195\textwidth}
    \centering
    \includegraphics[width=\textwidth]{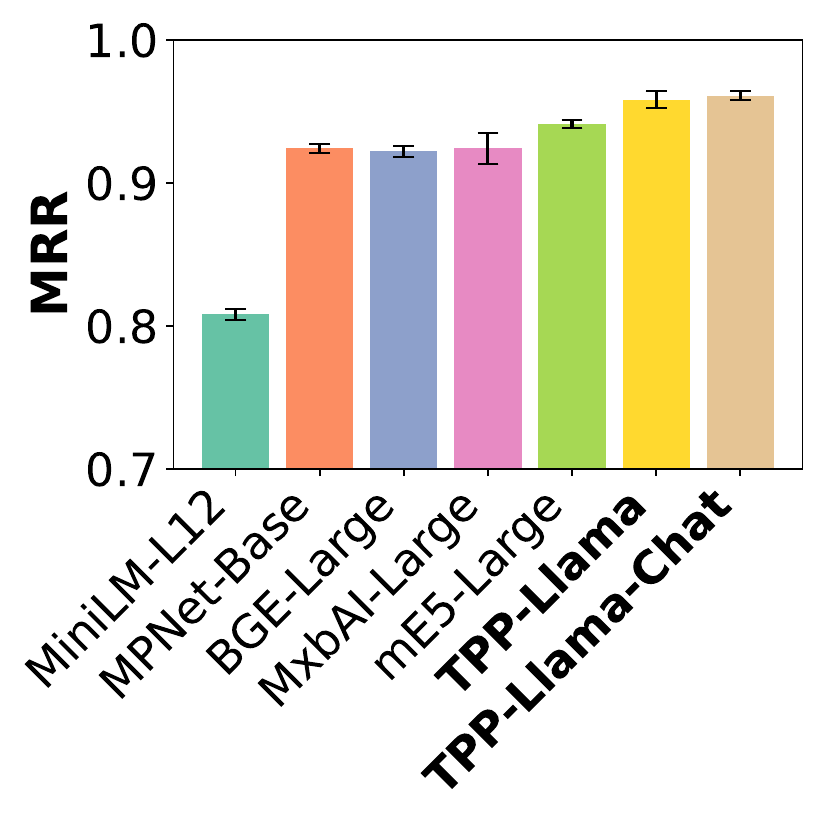}
    \caption{Chicago Crime}
\end{subfigure}
\hfill
\begin{subfigure}[b]{0.195\textwidth}
    \centering
    \includegraphics[width=\textwidth]{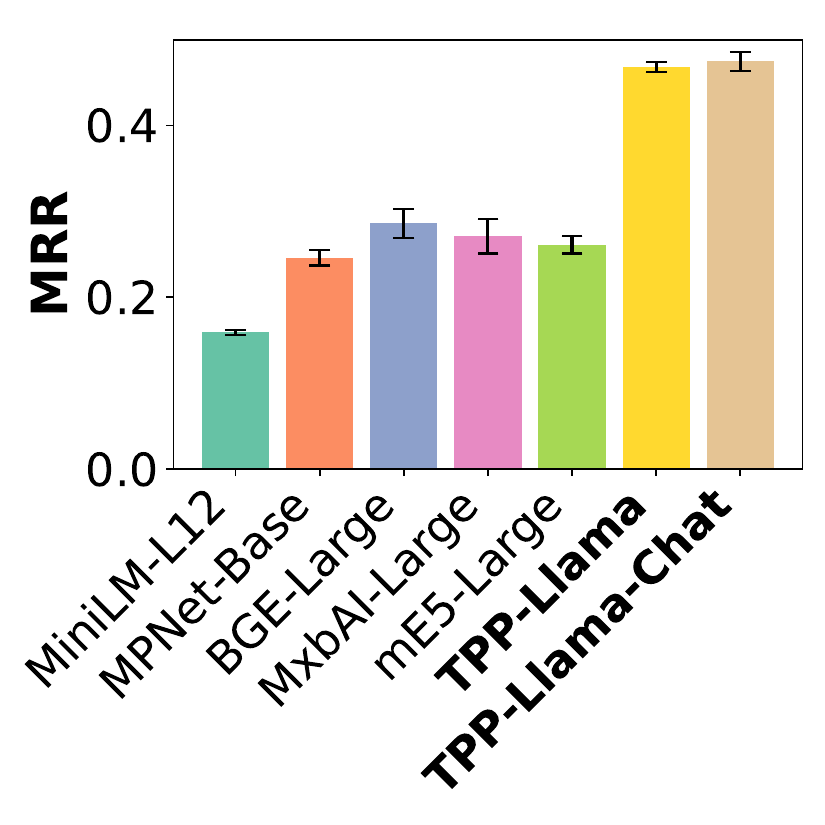}
    \caption{NYC Taxi Trip}
\end{subfigure}
\hfill
\begin{subfigure}[b]{0.195\textwidth}
    \centering
    \includegraphics[width=\textwidth]{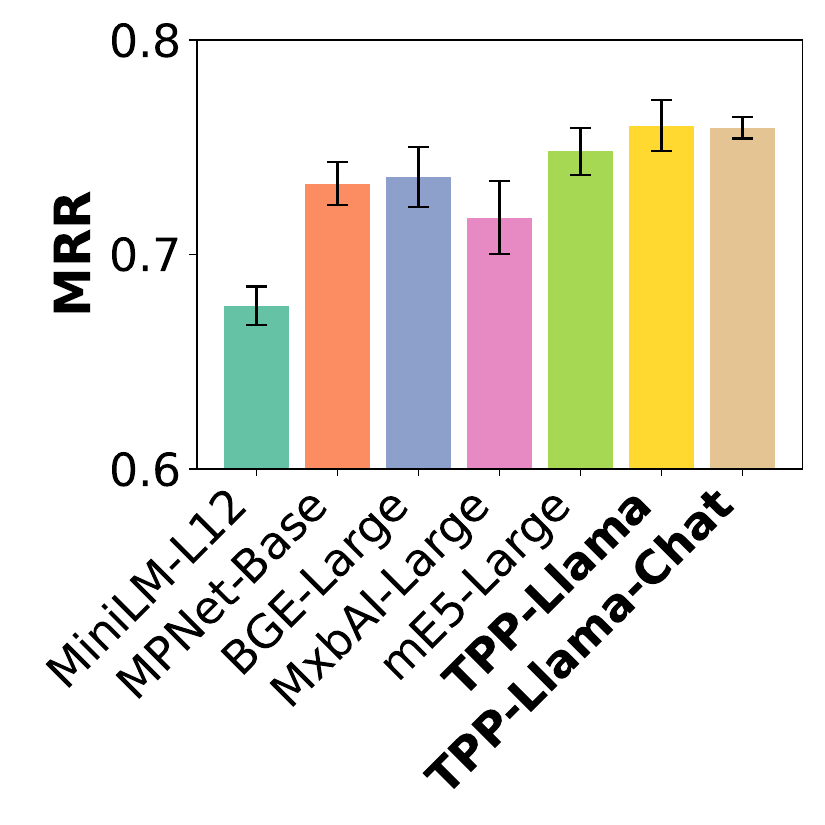}
    \caption{U.S. Earthquake}
\end{subfigure}
\hfill
\begin{subfigure}[b]{0.195\textwidth}
    \centering
    \includegraphics[width=\textwidth]{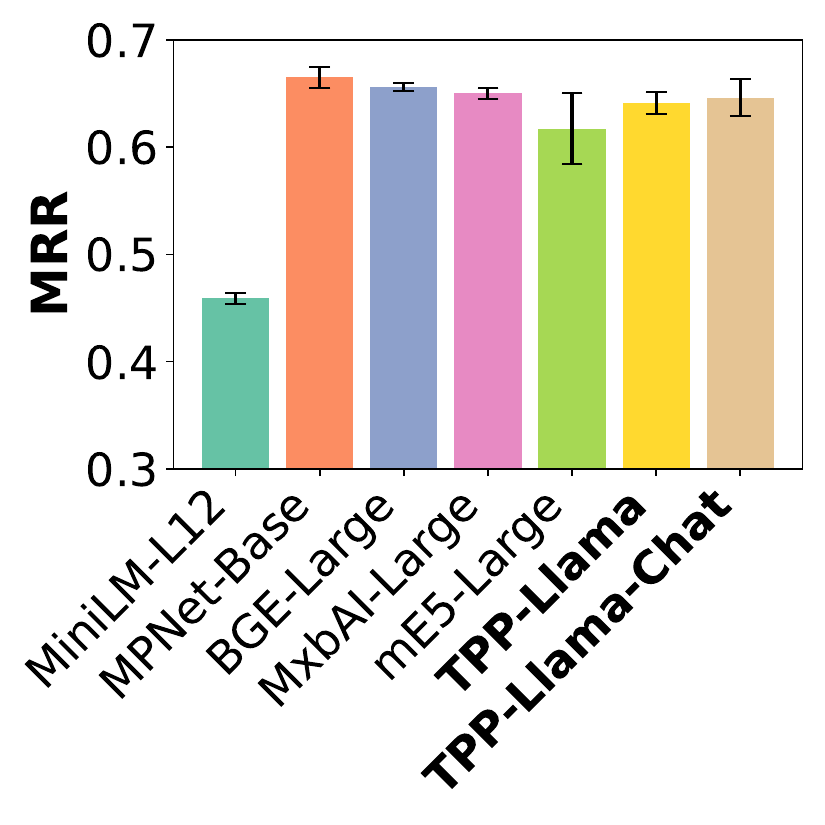}
    \caption{Amazon Review}
\end{subfigure}
\caption{Comparison of average MRRs with standard deviations on TESRBench in event sequence retrieval.}
\label{fig:mrrs}
\end{figure*}

\begin{figure*}[!h]
\centering
\begin{subfigure}[b]{0.195\textwidth}
    \centering
    \includegraphics[width=\textwidth]{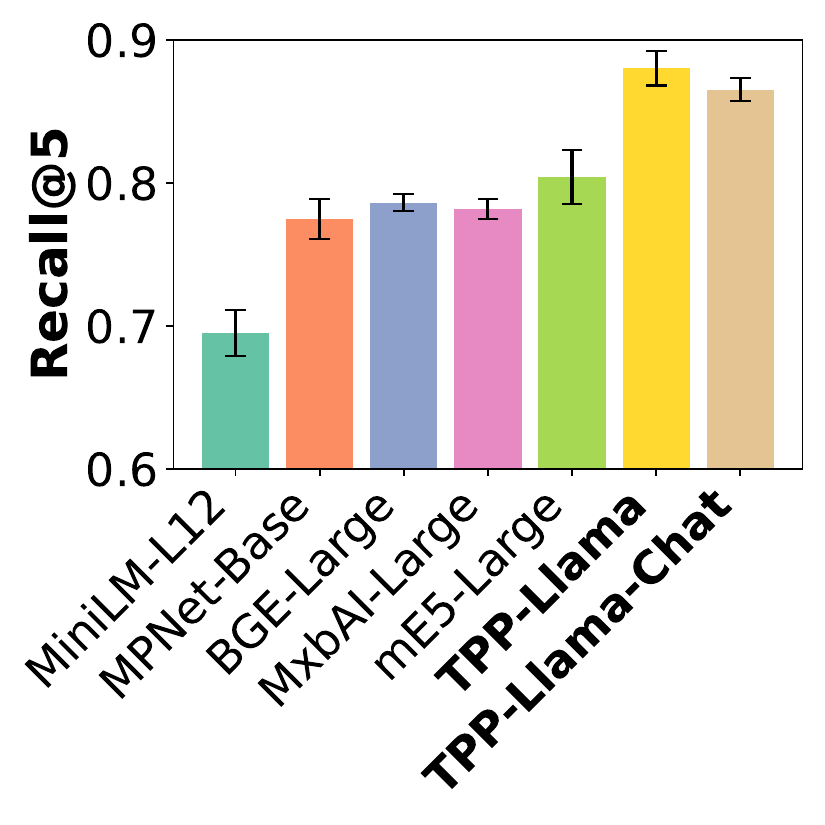}
    \caption{Stack Overflow}
\end{subfigure}
\hfill
\begin{subfigure}[b]{0.195\textwidth}
    \centering
    \includegraphics[width=\textwidth]{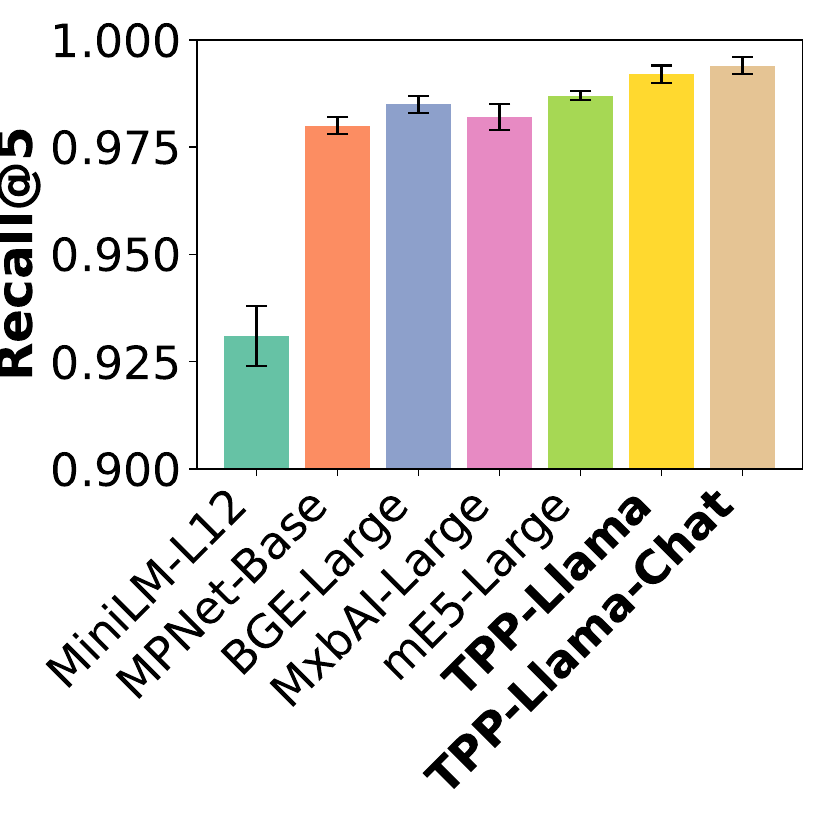}
    \caption{Chicago Crime}
\end{subfigure}
\hfill
\begin{subfigure}[b]{0.195\textwidth}
    \centering
    \includegraphics[width=\textwidth]{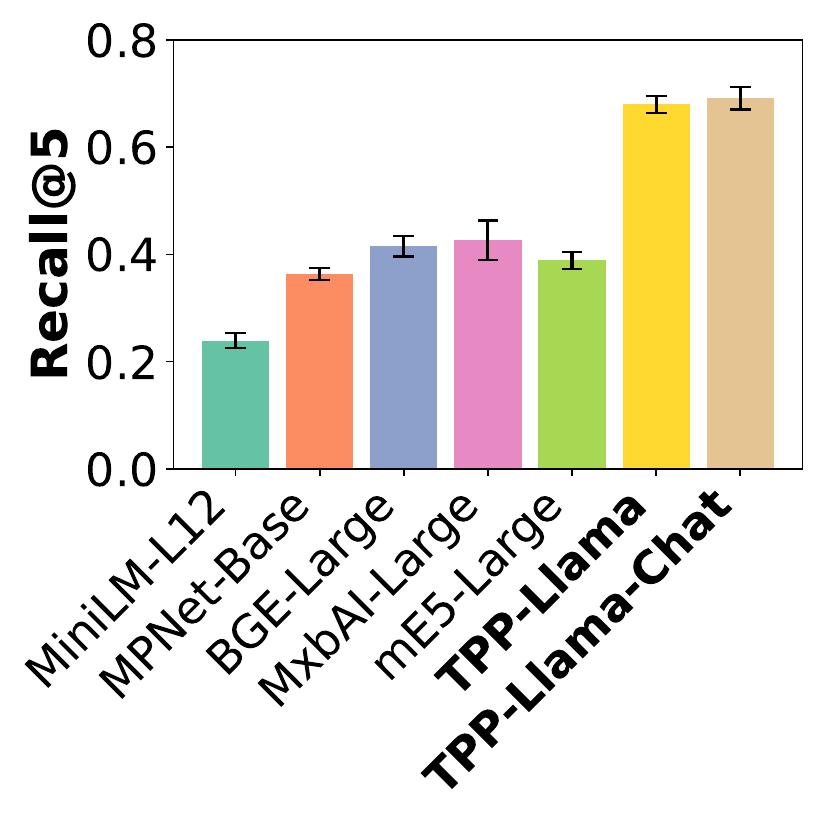}
    \caption{NYC Taxi Trip}
\end{subfigure}
\hfill
\begin{subfigure}[b]{0.195\textwidth}
    \centering
    \includegraphics[width=\textwidth]{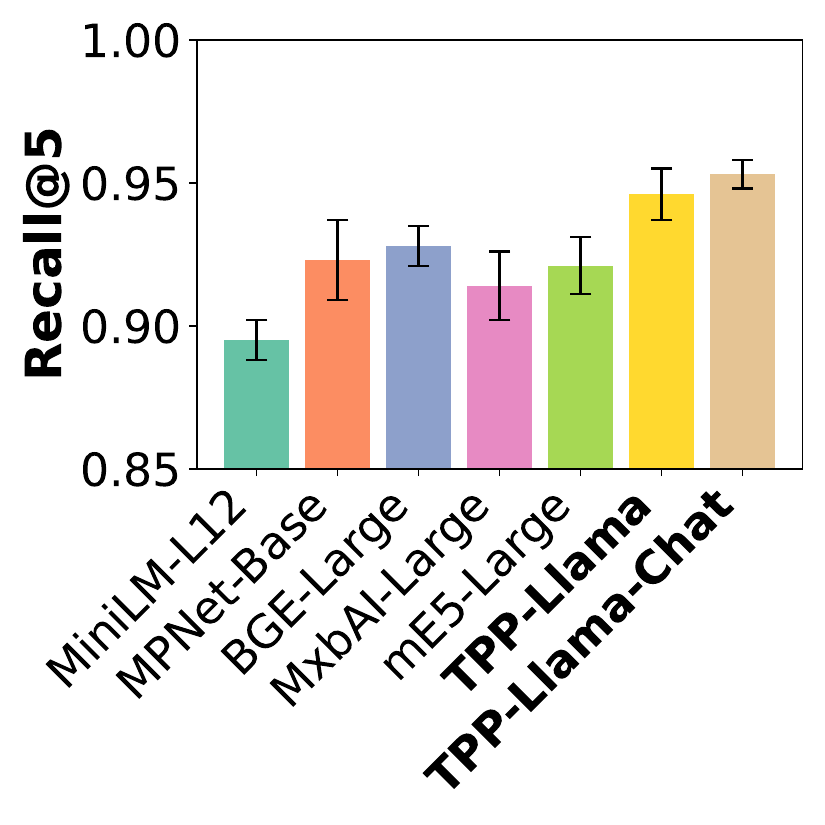}
    \caption{U.S. Earthquake}
\end{subfigure}
\hfill
\begin{subfigure}[b]{0.195\textwidth}
    \centering
    \includegraphics[width=\textwidth]{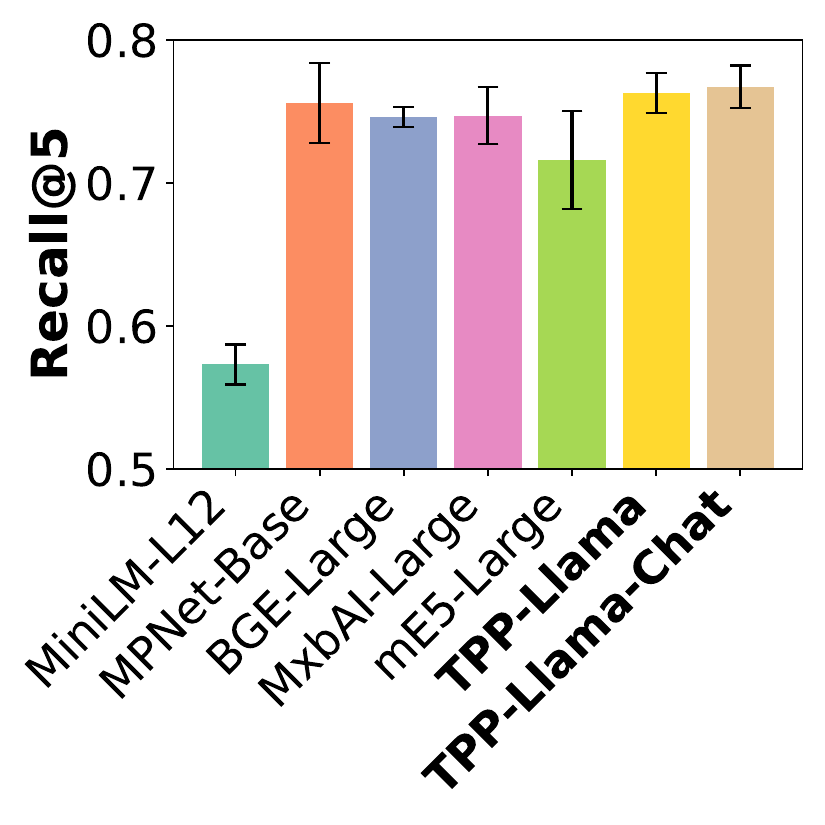}
    \caption{Amazon Review}
\end{subfigure}
\caption{Comparison of average Recall@5 with standard deviations on TESRBench in event sequence retrieval.}
\label{fig:realls-5}
\end{figure*}

The experimental results demonstrate the effectiveness of our proposed models compared to traditional text-based embedding models. As shown in Table \ref{tab:results}, along with Figures \ref{fig:mrrs} and \ref{fig:realls-5}, TPP-Llama and TPP-Llama-Chat consistently outperform the baselines across most datasets in terms of both MRR and Recall@5. TPP-Llama achieves the highest MRR on the Stack Overflow and U.S. Earthquake datasets, as well as the best Recall@5 on Stack Overflow, demonstrating its robustness across diverse event sequence retrieval tasks. TPP-Llama-Chat excels on the Chicago Crime and NYC Taxi Trip datasets, where it achieves top performance in both MRR and Recall@5. While MPNet-Base and BGE-Large provide competitive results on the Amazon Review dataset, the TPP-based models exhibit superior generalization across the majority of datasets. These results highlight the advantage of the temporal and event-type-aware design of TPP-Embedding, which effectively captures the structure and dependencies within event sequences compared to traditional models.

\subsection{Multi-Domain Results}

\begin{table}[!h]
\centering
\small
\begin{tabular}{cccc}
\toprule
\textbf{Model} & \textbf{MRR} & \textbf{Recall@5} \\
\midrule
MiniLM-L12 & 0.634 $\pm$ 0.007 & 0.795 $\pm$ 0.009 \\
MPNet-Base & 0.748 $\pm$ 0.003 &0.889 $\pm$ 0.007 \\
BGE-Large & 0.744 $\pm$ 0.010 & 0.888 $\pm$ 0.010 \\
MxbAI-Large & 0.744 $\pm$ 0.006 & 0.876 $\pm$ 0.010 \\
mE5-Large & 0.749 $\pm$ 0.013 & 0.888 $\pm$ 0.012 \\
\textbf{TPP-Llama} & \textbf{0.772} $\pm$ 0.009 & \underline{0.914} $\pm$ 0.008 \\
\textbf{TPP-Llama-Chat} & \underline{0.770} $\pm$ 0.005 & \textbf{0.919} $\pm$ 0.009 \\
\bottomrule
\end{tabular}
\caption{Comparison of average MRRs and Recall@5 with standard deviations on the multi-domain dataset.}
\label{tab:multi-domain-results}
\end{table}

In real-world applications, it is often necessary to retrieve event sequences that span different domains, requiring models to handle various event sequence types. Multi-domain retrieval refers to the ability of a model to effectively process and retrieve information across diverse datasets or domains simultaneously, rather than being specialized for a single domain. To simulate such settings, we created a multi-domain dataset by combining 30\% data from the above five datasets. As shown in Table \ref{tab:multi-domain-results}, the TPP-Llama and TPP-Llama-Chat models outperformed the baseline embedding models, achieving the highest MRRs and recall scores. The multi-domain experiments demonstrate the effectiveness of TPP-Embedding in handling diverse event sequences, highlighting their potential for use in applications requiring the retrieval of event sequences from multiple sources.

\subsection{Ablation Studies}

In this subsection, we perform ablation studies to evaluate the effects of various model configurations on event sequence retrieval performance.
\subsubsection{Embedding Inclusions}

We conduct an ablation study to assess the impact of using only temporal tokens or only type (textual) tokens on retrieval performance. As shown in Table \ref{tab:embedding-modes}, using only textual tokens achieves performance comparable to using all tokens on the Stack Overflow dataset. However, this approach leads to a significant performance drop on the U.S. Earthquake dataset, likely due to the nature of the datasets: Stack Overflow includes 25 event types, allowing the model to rely primarily on textual contents, whereas the U.S. Earthquake dataset contains only 3 event types, making temporal information essential for accurate retrieval.

\begin{table}[!h]
\centering
\small
\begin{tabular}{ccc}
\toprule
\textbf{Embeddings} & \textbf{StackOverflow} & \textbf{Earthquake} \\
\midrule
Temporal Tokens & 0.037 / 0.040 & 0.179 / 0.281 \\
Textual Tokens & 0.726 / \textbf{0.870} & 0.675 / 0.890 \\
\textbf{All Tokens} & \textbf{0.729} / 0.865 & \textbf{0.759} / \textbf{0.953} \\
\bottomrule
\end{tabular}
\caption{Comparison of average MRRs and Recall@5 of TPP-Llama-Chat with different embedding inclusions.}
\label{tab:embedding-modes}
\end{table}

\subsubsection{Hidden State Selections}

We evaluate the impact of different hidden state selections from the last hidden layer of the model for event sequences, specifically choosing only temporal tokens, a combination of temporal tokens and the last token of event type text tokens for each event, or all tokens. As shown in Table \ref{tab:hidden-state-modes}, using all tokens generally provides strong results, achieving the highest MRR on the StackOverflow dataset and the highest Recall@5 on the Earthquake dataset. While selecting temporal tokens and the last type tokens slightly improves MRR on the Earthquake dataset, using only temporal tokens lags behind both strategies on both datasets. Overall, choosing all tokens yields consistently good performance.

\begin{table}[!h]
\centering
\small
\begin{tabular}{ccc}
\toprule
\textbf{Hidden States} & \textbf{StackOverflow} & \textbf{Earthquake} \\
\midrule
Temporal Tokens & 0.718 / 0.862 & 0.754 / 0.939 \\
+ Last Type Tokens & 0.727 / \textbf{0.875} & \textbf{0.766} / \textbf{0.953} \\
\textbf{All Tokens} & \textbf{0.729} / 0.865 & 0.759 / \textbf{0.953} \\
\bottomrule
\end{tabular}
\caption{Comparison of average MRRs and Recall@5 of TPP-Llama-Chat with different hidden state selections.}
\label{tab:hidden-state-modes}
\end{table}

\subsubsection{Pooling Modes}

In experiments with different pooling modes as Table \ref{tab:pooling-modes}, we observe that the mean pooling method consistently performs well, achieving the highest MRR and Recall@5 on the StackOverflow dataset. However, for the Earthquake dataset, last token pooling \citep{muennighoff2022sgpt} slightly outperforms mean pooling. Max pooling shows competitive performance on the StackOverflow dataset but performs considerably worse on the Earthquake dataset. Overall, mean pooling offers a balanced performance, making it a reliable choice.

\begin{table}[!h]
\centering
\small
\begin{tabular}{ccc}
\toprule
\textbf{Pooling} & \textbf{StackOverflow} & \textbf{Earthquake} \\
\midrule
\textbf{Mean} & \textbf{0.729} / \textbf{0.865} & 0.759 / 0.953 \\
Max & 0.712 / 0.857 & 0.627 / 0.853 \\
Last Token & 0.728 / 0.848 & \textbf{0.772} / \textbf{0.960} \\
\bottomrule
\end{tabular}
\caption{Comparison of average MRRs and Recall@5 of TPP-Llama-Chat with different pooling modes.}
\label{tab:pooling-modes}
\end{table}

\subsubsection{Loss Functions}

To examine the impact of the loss function on retrieval performance, we replace the contrastive loss with a Mean Squared Error (MSE) loss, which optimizes cosine similarity to 1 for matched pairs. As shown in Table \ref{tab:loss-functions}, this substitution leads to a pronounced decline in both metrics across all datasets, emphasizing the pivotal role of contrastive loss in capturing subtle relationships between closely related event sequences. These results highlight the effectiveness of contrastive learning in enhancing retrieval accuracy.

\begin{table}[!h]
\centering
\small
\begin{tabular}{ccc}
\toprule
\textbf{Loss} & \textbf{StackOverflow} & \textbf{Earthquake} \\
\midrule
MSE & 0.020 / 0.016 & 0.020 / 0.015 \\
\textbf{Contrastive} & \textbf{0.729} / \textbf{0.865} & \textbf{0.759} / \textbf{0.953} \\
\bottomrule
\end{tabular}
\caption{Comparison of average MRRs and Recall@5 of TPP-Llama-Chat with different loss functions.}
\label{tab:loss-functions}
\end{table}

\section{Conclusion}

In this paper, we introduce TESRBench, a comprehensive benchmark for evaluating temporal event sequence retrieval, alongside TPP-Embedding, a novel model designed to integrate temporal and event-type-aware representations. TESRBench provides a diverse set of datasets with synthesized textual descriptions, offering a robust foundation for benchmarking models in this domain. Our proposed TPP-Embedding model combines temporal encoding and event text embedding with a large language model backbone, enabling it to effectively capture the structure and dependencies of temporal event sequences. Extensive experiments conducted on TESRBench demonstrate its superior performance compared to traditional text-based baselines, particularly in handling temporally complex, multi-type event sequences. Furthermore, multi-domain experiments underscore the flexibility and adaptability of our approach across diverse event domains. Together, TESRBench and TPP-Embedding represent a significant step forward in advancing research on temporal event sequence retrieval.

\section*{Limitations}

TESRBench, while providing a robust foundation for evaluating temporal event sequence retrieval, relies on synthesized textual descriptions generated by GPT-4o-mini, which may not fully capture the variability and complexity of real-world user-generated descriptions. A limitation of our TPP-Embedding model is its reliance on high-quality temporal and event-type data, which could pose challenges when dealing with noisy or incomplete event sequences encountered in real-world scenarios. Furthermore, while TPP-Embedding achieves strong retrieval performance, its dependence on large-scale language models can introduce computational latency on extremely large datasets, necessitating further optimization strategies. Finally, our current baselines are restricted to text-based methods, and future research could explore integrating recent time-context-aware sequential recommendation techniques \citep{li2020time,rashed2022context,tran2023attention,liu2024unirec} to further improve the retrieval of temporal event sequences from textual descriptions.

\section*{Ethical Considerations}

In constructing TESRBench, we acknowledge potential ethical concerns related to the use of synthesized textual descriptions and real-world event data. While the textual descriptions are generated objectively, they may still inadvertently reflect biases or limitations inherent in the data sources. For TPP-Embedding, its ability to retrieve temporal event sequences could be misused in privacy-sensitive applications, such as personal activity tracking. It is crucial to ensure that all data used for training and retrieval is anonymized and managed responsibly. Additionally, biases in training data, such as uneven representation of event types or domains, could result in biased retrieval outcomes. Future work should emphasize dataset curation and the implementation of bias mitigation strategies to minimize potential harms.

\bibliography{refs}

\begin{thebibliography}{39}
\providecommand{\natexlab}[1]{#1}

\bibitem[{Achiam et~al.(2023)Achiam, Adler, Agarwal, Ahmad, Akkaya, Aleman, Almeida, Altenschmidt, Altman, Anadkat et~al.}]{achiam2023gpt}
Josh Achiam, Steven Adler, Sandhini Agarwal, Lama Ahmad, Ilge Akkaya, Florencia~Leoni Aleman, Diogo Almeida, Janko Altenschmidt, Sam Altman, Shyamal Anadkat, et~al. 2023.
\newblock Gpt-4 technical report.
\newblock \emph{arXiv preprint arXiv:2303.08774}.

\bibitem[{{Chicago Police Department}(2024)}]{chicago2024crimes}
{Chicago Police Department}. 2024.
\newblock \href {https://data.cityofchicago.org/Public-Safety/Crimes-2001-to-Present/ijzp-q8t2} {Crimes - 2001 to present}.

\bibitem[{Dettmers et~al.(2022)Dettmers, Lewis, Belkada, and Zettlemoyer}]{dettmers2022llmint8}
Tim Dettmers, Mike Lewis, Younes Belkada, and Luke Zettlemoyer. 2022.
\newblock Llm.int8(): 8-bit matrix multiplication for transformers at scale.
\newblock \emph{arXiv preprint arXiv:2208.07339}.

\bibitem[{Dettmers et~al.(2024)Dettmers, Pagnoni, Holtzman, and Zettlemoyer}]{dettmers2024qlora}
Tim Dettmers, Artidoro Pagnoni, Ari Holtzman, and Luke Zettlemoyer. 2024.
\newblock Qlora: Efficient finetuning of quantized llms.
\newblock \emph{Advances in Neural Information Processing Systems}, 36.

\bibitem[{Gupta et~al.(2022)Gupta, Bedathur, and De}]{gupta2022learning}
Vinayak Gupta, Srikanta Bedathur, and Abir De. 2022.
\newblock Learning temporal point processes for efficient retrieval of continuous time event sequences.
\newblock In \emph{Proceedings of the AAAI Conference on Artificial Intelligence}, volume~36, pages 4005--4013.

\bibitem[{Gupta et~al.(2023)Gupta, Bedathur, and De}]{gupta2023retrieving}
Vinayak Gupta, Srikanta Bedathur, and Abir De. 2023.
\newblock Retrieving continuous time event sequences using neural temporal point processes with learnable hashing.
\newblock \emph{ACM Transactions on Intelligent Systems and Technology}.

\bibitem[{Henderson et~al.(2017)Henderson, Al-Rfou, Strope, Sung, Luk{\'a}cs, Guo, Kumar, Miklos, and Kurzweil}]{henderson2017efficient}
Matthew Henderson, Rami Al-Rfou, Brian Strope, Yun-Hsuan Sung, L{\'a}szl{\'o} Luk{\'a}cs, Ruiqi Guo, Sanjiv Kumar, Balint Miklos, and Ray Kurzweil. 2017.
\newblock Efficient natural language response suggestion for smart reply.
\newblock \emph{arXiv preprint arXiv:1705.00652}.

\bibitem[{Hu et~al.(2021)Hu, Shen, Wallis, Allen-Zhu, Li, Wang, Wang, and Chen}]{hu2021lora}
Edward~J Hu, Yelong Shen, Phillip Wallis, Zeyuan Allen-Zhu, Yuanzhi Li, Shean Wang, Lu~Wang, and Weizhu Chen. 2021.
\newblock Lora: Low-rank adaptation of large language models.
\newblock \emph{arXiv preprint arXiv:2106.09685}.

\bibitem[{Kashyap et~al.(2024)Kashyap, Nguyen, Schlegel, Winkler, Ng, and Poria}]{kashyap2024comprehensive}
Abhinav~Ramesh Kashyap, Thanh-Tung Nguyen, Viktor Schlegel, Stefan Winkler, See~Kiong Ng, and Soujanya Poria. 2024.
\newblock A comprehensive survey of sentence representations: From the bert epoch to the chatgpt era and beyond.
\newblock In \emph{Proceedings of the 18th Conference of the European Chapter of the Association for Computational Linguistics (Volume 1: Long Papers)}, pages 1738--1751.

\bibitem[{Lee et~al.(2024)Lee, Shakir, Koenig, and Lipp}]{emb2024mxbai}
Sean Lee, Aamir Shakir, Darius Koenig, and Julius Lipp. 2024.
\newblock \href {https://www.mixedbread.ai/blog/mxbai-embed-large-v1} {Open source strikes bread - new fluffy embeddings model}.

\bibitem[{Lhoest et~al.(2021)Lhoest, Del~Moral, Jernite, Thakur, Von~Platen, Patil, Chaumond, Drame, Plu, Tunstall et~al.}]{lhoest2021datasets}
Quentin Lhoest, Albert~Villanova Del~Moral, Yacine Jernite, Abhishek Thakur, Patrick Von~Platen, Suraj Patil, Julien Chaumond, Mariama Drame, Julien Plu, Lewis Tunstall, et~al. 2021.
\newblock Datasets: A community library for natural language processing.
\newblock \emph{arXiv preprint arXiv:2109.02846}.

\bibitem[{Li et~al.(2020)Li, Wang, and McAuley}]{li2020time}
Jiacheng Li, Yujie Wang, and Julian McAuley. 2020.
\newblock Time interval aware self-attention for sequential recommendation.
\newblock In \emph{Proceedings of the 13th international conference on web search and data mining}, pages 322--330.

\bibitem[{Li and Li(2023)}]{li2023angle}
Xianming Li and Jing Li. 2023.
\newblock Angle-optimized text embeddings.
\newblock \emph{arXiv preprint arXiv:2309.12871}.

\bibitem[{Lin et~al.(2022)Lin, Nogueira, and Yates}]{lin2022pretrained}
Jimmy Lin, Rodrigo Nogueira, and Andrew Yates. 2022.
\newblock \emph{Pretrained transformers for text ranking: Bert and beyond}.
\newblock Springer Nature.

\bibitem[{Liu et~al.(2024)Liu, Wang, and Feng}]{liu2024unirec}
Yang Liu, Yitong Wang, and Chenyue Feng. 2024.
\newblock Unirec: A dual enhancement of uniformity and frequency in sequential recommendations.
\newblock In \emph{Proceedings of the 33rd ACM International Conference on Information and Knowledge Management}, pages 1483--1492.

\bibitem[{Liu and Quan(2024)}]{liu2024tppllmm}
Zefang Liu and Yinzhu Quan. 2024.
\newblock Tpp-llm: Modeling temporal point processes by efficiently fine-tuning large language models.
\newblock \emph{arXiv preprint arXiv:2410.02062}.

\bibitem[{Loshchilov and Hutter(2017)}]{loshchilov2017decoupled}
Ilya Loshchilov and Frank Hutter. 2017.
\newblock Decoupled weight decay regularization.
\newblock \emph{arXiv preprint arXiv:1711.05101}.

\bibitem[{Mangrulkar et~al.(2022)Mangrulkar, Gugger, Debut, Belkada, Paul, and Bossan}]{mangrulkar2022peft}
Sourab Mangrulkar, Sylvain Gugger, Lysandre Debut, Younes Belkada, Sayak Paul, and B.~Bossan. 2022.
\newblock \href {https://github.com/huggingface/peft} {{PEFT}: State-of-the-art parameter-efficient fine-tuning methods}.

\bibitem[{Mei and Eisner(2017)}]{mei2017neural}
Hongyuan Mei and Jason~M Eisner. 2017.
\newblock The neural hawkes process: A neurally self-modulating multivariate point process.
\newblock \emph{Advances in neural information processing systems}, 30.

\bibitem[{Monroy-Hernandez(2014)}]{nyctaxi2024trips}
Andres Monroy-Hernandez. 2014.
\newblock \href {https://www.andresmh.com/nyctaxitrips} {{NYC} taxi trips}.

\bibitem[{Muennighoff(2022)}]{muennighoff2022sgpt}
Niklas Muennighoff. 2022.
\newblock Sgpt: Gpt sentence embeddings for semantic search.
\newblock \emph{arXiv preprint arXiv:2202.08904}.

\bibitem[{Ni et~al.(2019)Ni, Li, and McAuley}]{ni2019justifying}
Jianmo Ni, Jiacheng Li, and Julian McAuley. 2019.
\newblock Justifying recommendations using distantly-labeled reviews and fine-grained aspects.
\newblock In \emph{Proceedings of the 2019 conference on empirical methods in natural language processing and the 9th international joint conference on natural language processing (EMNLP-IJCNLP)}, pages 188--197.

\bibitem[{Paszke et~al.(2019)Paszke, Gross, Massa, Lerer, Bradbury, Chanan, Killeen, Lin, Gimelshein, Antiga et~al.}]{paszke2019pytorch}
Adam Paszke, Sam Gross, Francisco Massa, Adam Lerer, James Bradbury, Gregory Chanan, Trevor Killeen, Zeming Lin, Natalia Gimelshein, Luca Antiga, et~al. 2019.
\newblock Pytorch: An imperative style, high-performance deep learning library.
\newblock \emph{Advances in neural information processing systems}, 32.

\bibitem[{Rashed et~al.(2022)Rashed, Elsayed, and Schmidt-Thieme}]{rashed2022context}
Ahmed Rashed, Shereen Elsayed, and Lars Schmidt-Thieme. 2022.
\newblock Context and attribute-aware sequential recommendation via cross-attention.
\newblock In \emph{Proceedings of the 16th ACM Conference on Recommender Systems}, pages 71--80.

\bibitem[{Reimers and Gurevych(2019)}]{reimers2019sentence}
Nils Reimers and Iryna Gurevych. 2019.
\newblock Sentence-bert: Sentence embeddings using siamese bert-networks.
\newblock In \emph{Proceedings of the 2019 Conference on Empirical Methods in Natural Language Processing}.

\bibitem[{Shchur et~al.(2021)Shchur, T{\"u}rkmen, Januschowski, and G{\"u}nnemann}]{shchur2021neural}
Oleksandr Shchur, Ali~Caner T{\"u}rkmen, Tim Januschowski, and Stephan G{\"u}nnemann. 2021.
\newblock Neural temporal point processes: A review.
\newblock \emph{arXiv preprint arXiv:2104.03528}.

\bibitem[{Song et~al.(2020)Song, Tan, Qin, Lu, and Liu}]{song2020mpnet}
Kaitao Song, Xu~Tan, Tao Qin, Jianfeng Lu, and Tie-Yan Liu. 2020.
\newblock Mpnet: Masked and permuted pre-training for language understanding.
\newblock \emph{Advances in neural information processing systems}, 33:16857--16867.

\bibitem[{{Stack Exchange, Inc.}(2024)}]{stackexchange2024dump}
{Stack Exchange, Inc.} 2024.
\newblock \href {https://archive.org/details/stackexchange} {Stack exchange data dump}.

\bibitem[{Tran et~al.(2023)Tran, Salha-Galvan, Sguerra, and Hennequin}]{tran2023attention}
Viet~Anh Tran, Guillaume Salha-Galvan, Bruno Sguerra, and Romain Hennequin. 2023.
\newblock Attention mixtures for time-aware sequential recommendation.
\newblock In \emph{Proceedings of the 46th International ACM SIGIR Conference on Research and Development in Information Retrieval}, pages 1821--1826.

\bibitem[{{U.S. Geological Survey}(2024)}]{usgs2024earthquakes}
{U.S. Geological Survey}. 2024.
\newblock \href {https://earthquake.usgs.gov/earthquakes/search} {{USGS} earthquake catalog}.

\bibitem[{Vaswani et~al.(2017)Vaswani, Shazeer, Parmar, Uszkoreit, Jones, Gomez, Kaiser, and Polosukhin}]{vaswani2017attention}
Ashish Vaswani, Noam Shazeer, Niki Parmar, Jakob Uszkoreit, Llion Jones, Aidan~N Gomez, {\L}ukasz Kaiser, and Illia Polosukhin. 2017.
\newblock Attention is all you need.
\newblock \emph{Advances in neural information processing systems}, 30.

\bibitem[{Wang et~al.(2024)Wang, Yang, Huang, Yang, Majumder, and Wei}]{wang2024multilingual}
Liang Wang, Nan Yang, Xiaolong Huang, Linjun Yang, Rangan Majumder, and Furu Wei. 2024.
\newblock Multilingual e5 text embeddings: A technical report.
\newblock \emph{arXiv preprint arXiv:2402.05672}.

\bibitem[{Wang et~al.(2020)Wang, Wei, Dong, Bao, Yang, and Zhou}]{wang2020minilm}
Wenhui Wang, Furu Wei, Li~Dong, Hangbo Bao, Nan Yang, and Ming Zhou. 2020.
\newblock Minilm: Deep self-attention distillation for task-agnostic compression of pre-trained transformers.
\newblock \emph{Advances in Neural Information Processing Systems}, 33:5776--5788.

\bibitem[{Wolf et~al.(2020)Wolf, Debut, Sanh, Chaumond, Delangue, Moi, Cistac, Rault, Louf, Funtowicz et~al.}]{wolf2020transformers}
Thomas Wolf, Lysandre Debut, Victor Sanh, Julien Chaumond, Clement Delangue, Anthony Moi, Pierric Cistac, Tim Rault, R{\'e}mi Louf, Morgan Funtowicz, et~al. 2020.
\newblock Transformers: State-of-the-art natural language processing.
\newblock In \emph{Proceedings of the 2020 conference on empirical methods in natural language processing: system demonstrations}, pages 38--45.

\bibitem[{Xiao et~al.(2023)Xiao, Liu, Zhang, Muennighoff, Lian, and Nie}]{xiao2023cpack}
Shitao Xiao, Zheng Liu, Peitian Zhang, Niklas Muennighoff, Defu Lian, and Jian-Yun Nie. 2023.
\newblock C-pack: Packaged resources to advance general chinese embedding.
\newblock \emph{arXiv preprint arXiv:2309.07597}.

\bibitem[{Xue et~al.(2023)Xue, Shi, Chu, Wang, Zhou, Hao, Jiang, Pan, Xu, Zhang et~al.}]{xue2023easytpp}
Siqiao Xue, Xiaoming Shi, Zhixuan Chu, Yan Wang, Fan Zhou, Hongyan Hao, Caigao Jiang, Chen Pan, Yi~Xu, James~Y Zhang, et~al. 2023.
\newblock Easytpp: Towards open benchmarking the temporal point processes.
\newblock \emph{arXiv preprint arXiv:2307.08097}.

\bibitem[{Zhang et~al.(2024)Zhang, Zeng, Wang, and Lu}]{zhang2024tinyllama}
Peiyuan Zhang, Guangtao Zeng, Tianduo Wang, and Wei Lu. 2024.
\newblock Tinyllama: An open-source small language model.
\newblock \emph{arXiv preprint arXiv:2401.02385}.

\bibitem[{Zhang et~al.(2020)Zhang, Lipani, Kirnap, and Yilmaz}]{zhang2020self}
Qiang Zhang, Aldo Lipani, Omer Kirnap, and Emine Yilmaz. 2020.
\newblock Self-attentive hawkes process.
\newblock In \emph{International conference on machine learning}, pages 11183--11193. PMLR.

\bibitem[{Zuo et~al.(2020)Zuo, Jiang, Li, Zhao, and Zha}]{zuo2020transformer}
Simiao Zuo, Haoming Jiang, Zichong Li, Tuo Zhao, and Hongyuan Zha. 2020.
\newblock Transformer hawkes process.
\newblock In \emph{International conference on machine learning}, pages 11692--11702. PMLR.

\end{thebibliography}
\appendix
\begin{table*}[!h]
\centering
\small
\begin{tabular}{ccccccc}
\toprule
\textbf{Model (before FT)} & \textbf{StackOverflow} & \textbf{Crime} & \textbf{Taxi} & \textbf{Earthquake} & \textbf{Amazon} & \textbf{Multi-Domain} \\
\midrule
MiniLM-L12 & 0.091 / 0.123 & 0.071 / 0.111 & 0.028 / 0.024 & 0.037 / 0.043 & 0.142 / 0.200 & 0.154 / 0.208 \\
MPNet-Base & 0.068 / 0.087 & 0.027 / 0.020 & 0.022 / 0.017 & 0.031 / 0.027 & 0.068 / 0.071 & 0.102 / 0.127 \\
BGE-Large & 0.122 / 0.162 & 0.126 / 0.158 & 0.042 / 0.051 & 0.039 / 0.040 & 0.215 / 0.293 & 0.196 / 0.247 \\
MxbAI-Large & 0.085 / 0.102 & 0.091 / 0.134 & 0.039 / 0.037 & 0.038 / 0.043 & 0.174 / 0.227 & 0.170 / 0.221 \\
mE5-Large & 0.065 / 0.078 & 0.078 / 0.087 & 0.028 / 0.024 & 0.037 / 0.040 & 0.142 / 0.187 & 0.145 / 0.191 \\
\textbf{TPP-Llama} & 0.022 / 0.021 & 0.019 / 0.020 & 0.020 / 0.014 & 0.022 / 0.020 & 0.033 / 0.027 & 0.025 / 0.030 \\
\textbf{TPP-Llama-Chat} & 0.020 / 0.015 & 0.018 / 0.012 & 0.019 / 0.014 & 0.023 / 0.020 & 0.033 / 0.031 & 0.021 / 0.017 \\
\bottomrule
\end{tabular}
\caption{Comparison of MRRs and Recall@5 on TESRBench in event sequence retrieval before fine-tuning.}
\label{tab:mrrs-recalls-no-ft}
\end{table*}

\begin{table*}[!h]
\centering
\small
\begin{tabular}{cccccc}
\toprule
\textbf{Model} & \textbf{StackOverflow} & \textbf{Crime} & \textbf{Taxi} & \textbf{Earthquake} & \textbf{Amazon} \\
\midrule
MiniLM-L12 & 0.501 $\pm$ 0.009 & 0.808 $\pm$ 0.004 & 0.159 $\pm$ 0.003 & 0.676 $\pm$ 0.009 & 0.459 $\pm$ 0.005 \\
MPNet-Base & 0.620 $\pm$ 0.007 & 0.924 $\pm$ 0.003 & 0.246 $\pm$ 0.009 & 0.733 $\pm$ 0.010 & \textbf{0.665} $\pm$ 0.010 \\
BGE-Large & 0.632 $\pm$ 0.007 & 0.922 $\pm$ 0.004 & 0.286 $\pm$ 0.017 & 0.736 $\pm$ 0.014 & \underline{0.656} $\pm$ 0.004 \\
MxbAI-Large & 0.627 $\pm$ 0.013 & 0.924 $\pm$ 0.011 & 0.271 $\pm$ 0.020 & 0.717 $\pm$ 0.017 & 0.650 $\pm$ 0.005 \\
mE5-Large & 0.658 $\pm$ 0.012 & 0.941 $\pm$ 0.003 & 0.261 $\pm$ 0.010 & 0.748 $\pm$ 0.011 & 0.617 $\pm$ 0.033 \\
\textbf{TPP-Llama} & \textbf{0.741} $\pm$ 0.006 & \underline{0.958} $\pm$ 0.006 & \underline{0.468} $\pm$ 0.006 & \textbf{0.760} $\pm$ 0.012 & 0.641 $\pm$ 0.010 \\
\textbf{TPP-Llama-Chat} & \underline{0.729} $\pm$ 0.008 & \textbf{0.961} $\pm$ 0.003 & \textbf{0.475} $\pm$ 0.011 & \underline{0.759} $\pm$ 0.005 & 0.646 $\pm$ 0.017 \\
\bottomrule
\end{tabular}
\caption{Comparison of average MRRs with standard deviations on TESRBench in event sequence retrieval.}
\label{tab:mrrs}
\end{table*}

\begin{table*}[!h]
\centering
\small
\begin{tabular}{cccccc}
\toprule
\textbf{Model} & \textbf{StackOverflow} & \textbf{Crime} & \textbf{Taxi} & \textbf{Earthquake} & \textbf{Amazon} \\
\midrule
MiniLM-L12 & 0.695 $\pm$ 0.016 & 0.931 $\pm$ 0.007 & 0.239 $\pm$ 0.014 & 0.895 $\pm$ 0.007 & 0.573 $\pm$ 0.014 \\
MPNet-Base & 0.775 $\pm$ 0.014 & 0.980 $\pm$ 0.002 & 0.364 $\pm$ 0.011 & 0.923 $\pm$ 0.014 & 0.756 $\pm$ 0.028 \\
BGE-Large & 0.786 $\pm$ 0.006 & 0.985 $\pm$ 0.002 & 0.415 $\pm$ 0.019 & 0.928 $\pm$ 0.007 & 0.746 $\pm$ 0.007 \\
MxbAI-Large & 0.782 $\pm$ 0.007 & 0.982 $\pm$ 0.003 & 0.426 $\pm$ 0.037 & 0.914 $\pm$ 0.012 & 0.747 $\pm$ 0.020 \\
mE5-Large & 0.804 $\pm$ 0.019 & 0.987 $\pm$ 0.001 & 0.389 $\pm$ 0.016 & 0.921 $\pm$ 0.010 & 0.716 $\pm$ 0.034 \\
\textbf{TPP-Llama} & \textbf{0.880} $\pm$ 0.012 & \underline{0.992} $\pm$ 0.002 & \underline{0.680} $\pm$ 0.016 & \underline{0.946} $\pm$ 0.009 & \underline{0.763} $\pm$ 0.014 \\
\textbf{TPP-Llama-Chat} & \underline{0.865} $\pm$ 0.008 & \textbf{0.994} $\pm$ 0.002 & \textbf{0.691} $\pm$ 0.021 & \textbf{0.953} $\pm$ 0.005 & \textbf{0.767} $\pm$ 0.015 \\
\bottomrule
\end{tabular}
\caption{Comparison of average Recall@5 with standard deviations on TESRBench in event sequence retrieval.}
\label{tab:recalls-5}
\end{table*}
\section{Data Examples}
\label{sec:data-examples}

This appendix presents selected examples of event sequences from the validation sets in TESRBench, along with their corresponding descriptions, as shown in Table \ref{tab:data-examples}. These descriptions highlight key temporal patterns and provide context for the diversity of events and their occurrences across the benchmark's datasets.

\section{More Experimental Setup}
\label{sec:more-setup}

Our experiments were conducted using several key Python libraries, including \texttt{pytorch} \citep{paszke2019pytorch} for deep learning, \texttt{transformers} \citep{wolf2020transformers} for working with pre-trained language models, \texttt{sentence-transformers} \citep{reimers2019sentence} for embedding and retrieval tasks, \texttt{datasets} \citep{lhoest2021datasets} for data handling, \texttt{peft} \citep{mangrulkar2022peft} for parameter-efficient fine-tuning, and \texttt{bitsandbytes} \citep{dettmers2022llmint8} for model quantization.

\section{More Experimental Results}
\label{sec:more-results}

In this appendix, we provide additional experimental results to further analyze the performance of our models, both before and after fine-tuning.

\subsection{Experimental Results before Fine-Tuning}

The performance of all models before fine-tuning in Table \ref{tab:mrrs-recalls-no-ft} shows significantly lower retrieval accuracy, with the highest MRR achieved by BGE-Large across most datasets. TPP-based models perform poorly at this stage, due to unaligned temporal and textual spaces, indicating the necessity of fine-tuning to improve event sequence retrieval.

\subsection{More Experimental Results after Fine-Tuning}

The results after fine-tuning demonstrate a significant improvement across all models, as shown in Tables \ref{tab:mrrs} and \ref{tab:recalls-5}. TPP-Llama and TPP-Llama-Chat consistently outperform baseline models, especially in MRR and Recall@5 across most datasets. These improvements highlight the importance of fine-tuning in enhancing their retrieval capabilities.

\begin{table*}[!h]
\centering
\small
\begin{tabular}{p{.15\linewidth}p{.25\linewidth}p{.5\linewidth}}
\toprule
\textbf{Dataset} & \textbf{Event Sequence} & \textbf{Description} \\
\midrule
Stack Overflow &
0.00,Nice Question\newline
0.57,Good Answer\newline
0.66,Popular Question\newline
0.83,Famous Question\newline
0.89,Nice Question\newline
2.25,Popular Question\newline
2.32,Necromancer\newline
2.43,Notable Question\newline
3.00,Great Question\newline
3.40,Good Question\newline
3.64,Popular Question\newline
3.69,Notable Question\newline
4.00,Notable Question\newline
\dots\newline
23.41,Popular Question
& The sequence of badges earned by the user on Stack Overflow shows a progression of achievements over time, with certain types of badges appearing more frequently at different intervals. Initially, badges such as "Nice Question" and "Good Answer" are earned in the first few months, followed by a cluster of "Popular Question" and "Notable Question" badges around the two to four-month mark. As time progresses, there is a notable increase in the frequency of "Famous Question" and "Popular Question" badges, particularly in the later months, indicating a trend toward these types of achievements. The sequence also includes several clusters of "Nice Question" badges, particularly around the seven-month mark and again towards the end of the timeline. Overall, the user's badge accumulation reflects a gradual increase in the variety and frequency of achievements over the course of the observed period. \\
\midrule
Chicago Crime
& 0.00,Battery\newline
0.26,Battery\newline
0.52,Theft\newline
0.69,Motor Vehicle Theft\newline
0.74,Assault\newline
0.99,Motor Vehicle Theft\newline
0.99,Motor Vehicle Theft\newline
1.08,Criminal Sexual Assault\newline
\dots\newline
23.72,Deceptive Practice
& The sequence of crime incidents shows a notable clustering of certain crime types over time, particularly motor vehicle thefts, which appear frequently throughout the timeline, especially in the earlier months. Battery incidents are also prevalent, occurring multiple times in the first half of the sequence. Other offenses such as robbery and criminal damage emerge at various intervals, with some clustering in the middle to later months. Overall, there is a trend of increasing diversity in crime types as the timeline progresses, with a gradual rise in the frequency of theft-related incidents towards the end. \\
\midrule
NYC Taxi Trip &
0.00,Manhattan Pickup\newline
0.19,Manhattan Dropoff\newline
0.24,Manhattan Pickup\newline
0.68,Manhattan Dropoff\newline
0.73,Manhattan Pickup\newline
0.99,Manhattan Dropoff\newline
1.13,Manhattan Pickup\newline
1.43,Manhattan Dropoff\newline
1.45,Manhattan Pickup\newline
1.54,Manhattan Dropoff\newline
\dots\newline
31.87,Brooklyn Dropoff
& The sequence of taxi trips primarily consists of pickups and dropoffs occurring in Manhattan, with a notable concentration of events in the first few hours. Early in the sequence, the driver consistently alternates between pickups and dropoffs, with a high frequency of trips. As the sequence progresses, there are brief periods where trips shift to Queens and Brooklyn, particularly after a long duration of Manhattan trips. The latter part of the sequence shows a gradual transition to more pickups and dropoffs in Brooklyn, indicating a shift in location focus. Overall, the events are clustered closely together in time, with significant activity in the first half of the sequence before expanding to other boroughs. \\
\midrule
U.S. Earthquake &
0.00,Medium\newline
0.66,Large\newline
0.72,Large\newline
0.99,Large\newline
1.07,Large\newline
1.08,Large\newline
1.67,Large
& The sequence of earthquake events begins with a medium magnitude event, followed closely by a series of large magnitude events occurring within a short time frame. The large events cluster together, with multiple occurrences happening within the first two days. This indicates a trend of increasing magnitude shortly after the initial medium event, with the majority of the large events occurring in rapid succession. \\
\midrule
Amazon Review &
0.00,Books\newline
0.14,Sports and Outdoors\newline
0.14,Books\newline
0.29,Books\newline
0.43,Books\newline
0.57,Books\newline
1.00,Books\newline
1.14,Books\newline
\dots\newline
25.29,Books
& The sequence of product reviews shows a predominant focus on the "Books" category, which appears consistently throughout the timeline, especially in the initial weeks. Other categories such as "Pet Supplies" and "Grocery and Gourmet Food" emerge intermittently, often clustering around specific weeks, particularly in the middle and later parts of the sequence. "Clothing Shoes and Jewelry" and "Movies and TV" also appear, but less frequently, with some clustering noted in the later weeks. Overall, there is a clear trend of sustained interest in "Books," with other categories appearing in a more sporadic manner. \\
\bottomrule
\end{tabular}
\caption{Event sequence examples with their descriptions from the validation sets of TESRBench.}
\label{tab:data-examples}
\end{table*}
\end{document}